\pdfoutput=1

\documentclass[11pt]{article}

\usepackage{naacl2021}

\usepackage{times}
\usepackage{latexsym}

\usepackage[T1]{fontenc}

\usepackage[utf8]{inputenc}

\usepackage{microtype}

\usepackage{graphicx}
\usepackage{subfigure}
\usepackage{xfrac}
\usepackage{colortbl}
\usepackage{booktabs}  
\usepackage{amsfonts}  
\usepackage{nicefrac}
\usepackage{latexsym}
\usepackage{multirow}
\usepackage{amsmath}
\usepackage{tabularx}

\newcommand*{\affmark}[1][*]{\textsuperscript{#1}}
\newcommand*{\email}[1]{\texttt{#1}}

\newcommand{\Ours}{QMSum}

\definecolor{topic}{RGB}{102,204,0}
\definecolor{speaker}{RGB}{255,102,102}
\definecolor{subtopic}{RGB}{204,0,204}

\title{\Ours{}: A New Benchmark for Query-based Multi-domain \\ Meeting Summarization}

\author{
Ming Zhong\thanks{\ \  These two authors contributed equally. The order of authorship decided by the flip of a coin.}\affmark[$*\dagger$] 
\quad Da Yin\affmark[$*\clubsuit$]
\quad Tao Yu\affmark[$\ddagger$]
\quad Ahmad Zaidi\affmark[$\ddagger$]\\
\bf \quad Mutethia Mutuma\affmark[$\ddagger$] 
\bf \quad Rahul Jha\affmark[$\P$]
\bf \quad Ahmed Hassan Awadallah\affmark[$\P$]\\
\bf \quad Asli Celikyilmaz\affmark[$\P$]
\bf \quad Yang Liu\affmark[$\S$]
\bf \quad Xipeng Qiu\affmark[$\dagger$] 
\bf \quad Dragomir Radev\affmark[$\ddagger$]
\\
{\affmark[$\dagger$]Fudan University} \quad
{\affmark[$\clubsuit$]University of California, Los Angeles} \quad
{\affmark[$\ddagger$]Yale University}\\
{\affmark[$\P$]Microsoft Research} \quad {\affmark[$\S$]Microsoft Cognitive Services Research}\\
\email{mzhong18@fudan.edu.cn} \quad \quad \email{da.yin@cs.ucla.edu} \\
\email{\{tao.yu,\,dragomir.radev\}@yale.edu}
}

\begin{document}

\maketitle

\begin{abstract}
Meetings are a key component of human collaboration. As increasing numbers of meetings are recorded and transcribed, meeting summaries have become essential to remind those who may or may not have attended the meetings about the key decisions made and the tasks to be completed. However, it is hard to create a single short summary that covers all the content of a long meeting involving multiple people and topics. In order to satisfy the needs of different types of users, we define a new query-based multi-domain meeting summarization task, where models have to select and summarize relevant spans of meetings in response to a query, and we introduce QMSum, a new benchmark for this task. QMSum consists of 1,808 query-summary pairs over 232 meetings in multiple domains. Besides, we investigate a \textit{locate-then-summarize} method and evaluate a set of strong summarization baselines on the task. Experimental results and manual analysis reveal that QMSum presents significant challenges in long meeting summarization for future research. Dataset is available at \url{https://github.com/Yale-LILY/QMSum}.

\end{abstract}

\section{Introduction}
Meetings remain the go-to tool for collaboration, with 11 million meetings taking place each day in the USA and employees spending six hours a week, on average, in meetings \cite{doi:10.1177/0963721418776307}. 
The emerging landscape of remote work is making meetings even more important and simultaneously taking a toll on our productivity and well-being~\cite{msft_study19}.
The proliferation of meetings makes it hard to stay on top of this sheer volume of information and increases the need for automated methods for accessing key information exchanged during them. 
Meeting summarization~\cite{wang-cardie-2013-domain,shang-etal-2018-unsupervised,li-etal-2019-keep,zhu-etal-2020-hierarchical} is a task where summarization models are leveraged to generate summaries of entire meetings based on meeting transcripts. 
The resulting summaries distill the core contents of a meeting that helps people efficiently catch up to meetings. 

\begin{figure}[t]
    \centering
    \includegraphics[width=\linewidth, trim=30 20 40 15, clip]{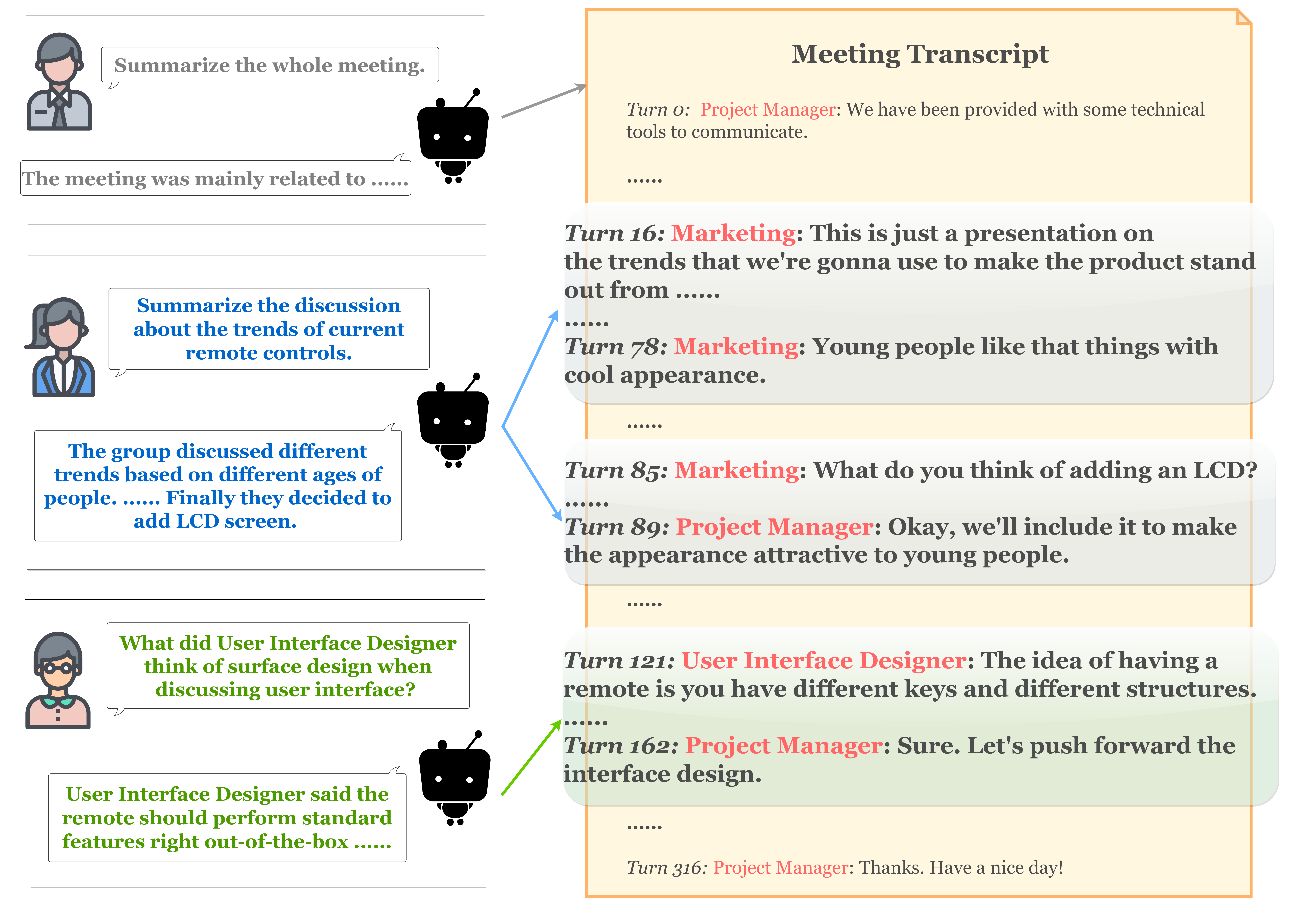}
    \caption{Examples of query-based meeting summarization task. Users are interested in different facets of the meeting. In this task, a model is required to summarize the contents that users are interested in and query.}
    \label{fig:intro_fig}
    \vspace{-2mm}
\end{figure}

Most existing work and datasets on meeting summarization~\cite{janin2003icsi,carletta2005ami} pose the problem as a single document summarization task where a single summary is generated for the whole meeting. 
Unlike news articles where people may be satisfied with a high-level summary, they are more likely to seek more detailed information when it comes to meeting summaries such as topics ~\cite{li-etal-2019-keep}, opinions, actions, and decisions~\cite{wang-cardie-2013-domain}. This poses the question of \emph{whether a single paragraph is enough to summarize the content of an entire meeting?}

Figure~\ref{fig:intro_fig} shows an example of a meeting about ``remote control design''. The discussions in the meeting are multi-faceted and hence different users might be interested in different facets. For example, someone may be interested in learning about the new trends that may lead to the new product standing out, while others may be more interested in what other attendees thought about different elements of the design. 
It is challenging to compress or compose a short summary that contains all the salient information. 
Alternatively, summarization systems should adopt a more flexible and interactive approach that allows people to express their interests and caters to their diverse intents when generating summaries~\cite{Dang2005OverviewOD,dang-2006-duc,litvak-vanetik-2017-query,baumel2018query}. 

With comprehensive consideration of the multi-granularity meeting contents, we propose a new task, query-based meeting summarization. To enable research in this area, we also create a high-quality multi-domain summarization dataset. In this task, as shown in Figure~\ref{fig:intro_fig}, given a query and a meeting transcript, a model is required to generate the corresponding summary. The query-based approach is a flexible setup that enables the system to satisfy different intents and different levels of granularity.
Besides the annotated queries and corresponding gold summaries at different levels of granularity, our new dataset contains a rich set of annotations that include the main topics of each meeting and the ranges of relevant text spans for the annotated topics and each query. 
We adopt a hierarchical annotation structure that could not only assist people to find information faster, but also strengthen the models' summarization capacity.

In this paper, we employ a two-stage meeting summarization approach: \textit{locate-then-summarize}. Specifically, given a query, a model called \textit{Locator} is used to locate the relevant utterances in the meeting transcripts, and then these extracted spans are used as an input to another model called \textit{Summarizer} to generate a query-based summary. We present and evaluate several strong baselines based on state-of-the-art summarization models on QMSum. Our results and analysis from different perspectives reveal that the existing models struggle in solving this task, highlighting the challenges the models face when generating query-based meeting summaries. 
We are releasing our dataset and baselines to support additional research in query-focused meeting summarization.

Overall, our contributions are listed as follows: 
1) We propose a new task, query-based multi-domain meeting summarization, and build a new benchmark QMSum with a hierarchical annotation structure. 2) We design a \textit{locate-then-summarize} model and conduct comprehensive experiments on its strong variants and different training settings. 3) By human evaluation, we further pose the challenges of the new task, including the impact of different query types and factuality errors.

\section{Related Work}

\subsection{Text Summarization}
Most prior work in text summarization~\cite{rush2015neural,chopra2016abstractive,nallapati2016abstractive,see2017get,celikyilmaz2018deep,chen-bansal-2018-fast,zhong2019searching,xu-durrett-2019-neural,liu2019text,lebanoff2019scoring,cho2019improving,DBLP:conf/acl/ZhongLCWQH20,DBLP:conf/acl/WangLZQH20,xu2019discourse,jia2020neural} investigate how to generate better summaries on news article data, such as CNN/DailyMail~\cite{hermann2015teaching}, Newsroom~\cite{grusky-etal-2018-newsroom}, etc. Scientific paper summarization is another important branch~\cite{cohan2018discourse,yasunaga2019scisummnet,an2021enhancing}. Our paper mainly focuses on meeting summarization, a more challenging task compared to news summarization. 
With the burst of demand for meeting summarization, this task attracts more and more interests from academia~\cite{wang-cardie-2013-domain,oya-etal-2014-template,shang-etal-2018-unsupervised,zhu-etal-2020-hierarchical} and becomes an emerging branch of text summarization area. 

\subsection{Query-based Summarization}
Query-based summarization aims to generate a brief summary according to a source document and a given query. There are works studying this task~\cite{daume-iii-marcu-2006-bayesian,otterbacher2009biased,wang2016sentence,litvak-vanetik-2017-query,nema-etal-2017-diversity,baumel2018query,ishigaki2020neural,kulkarni2020aquamuse,laskar2020query}. However, the models focus on news~\cite{Dang2005OverviewOD,dang-2006-duc}, debate~\cite{nema-etal-2017-diversity}, and Wikipedia~\cite{DBLP:journals/corr/abs-1911-03324}. Meeting is also a genre of discourses where query-based summarization could be applied, but to our best knowledge, there are no works studying this direction.

\subsection{Meeting Summarization}
Meeting summarization has attracted a lot of interest recently~\cite{chen2012integrating,wang-cardie-2013-domain,mehdad-etal-2013-abstractive,oya-etal-2014-template,shang-etal-2018-unsupervised,li-etal-2019-keep,zhu-etal-2020-hierarchical,koay-etal-2020-domain}.
Specifically, \citet{mehdad-etal-2013-abstractive} leverage entailment graphs and ranking strategy to generate meeting summaries. \citet{wang-cardie-2013-domain} attempt to make use of decisions, action items and progress to generate the whole meeting summaries. \citet{oya-etal-2014-template} leverages the relationship between summaries and the meeting transcripts to extract templates and generate summaries with the guidance of the templates.
\citet{shang-etal-2018-unsupervised} utilize multi-sentence compression techniques to generate summaries under an unsupervised setting.
\citet{li-etal-2019-keep} attempt to incorporate multi-modal information to facilitate the meeting summarization. \citet{zhu-etal-2020-hierarchical} propose a model which builds a hierarchical structure on word-level and turn-level information and uses news summary data to alleviate the inadequacy of meeting data. 

Unlike previous works, instead of merely generating summaries for the complete meeting, we propose a novel task where we focus on summarizing multi-granularity contents which cater to different people's need for the entire meetings, and help people comprehensively understand meetings.

\newcommand{\tabincell}[2]{\begin{tabular}{@{}#1@{}}#2\end{tabular}}

\section{Data Construction}
\label{data_construction}
In this section, we show how we collected meeting data from three different domains: academic meetings, product meetings, and committee meetings. In addition, we show how we annotated the three types of meeting data while ensuring annotation quality for query-based meeting summarization.

\subsection{Data Collection}
We introduce the three types of meetings that we used to annotate query-summary pairs. 

\paragraph{Product Meetings} AMI\footnote{\url{http://groups.inf.ed.ac.uk/ami/download/}}~\cite{carletta2005ami} is a dataset of meetings about product design in an industrial setting. It consists of 137 meetings about how to design a new remote control, from kick-off to completion over the course of a day.
It contains meeting transcripts and their corresponding meeting summaries.

\paragraph{Academic Meetings} ICSI\footnote{\url{http://groups.inf.ed.ac.uk/ami/icsi/index.shtml}}~\cite{janin2003icsi} dataset is an academic meeting dataset composed of 59 weekly group meetings at International Computer Science Institute (ICSI) in Berkeley, and their summaries. Different from AMI, the contents of ICSI meetings are specific to the discussions about research among students. 

\paragraph{Committee Meetings} Parliamentary committee meeting is another important domain of meetings. These meetings focus on the formal discussions on a wide range of issues (e.g., the reform of the education system, public health, etc.) Also, committee meetings are publicly available, which enables us to access large quantities of meetings. We include 25 committee meetings of the Welsh Parliament\footnote{\url{https://record.assembly.wales/}} and 11 from the Parliament of Canada\footnote{\url{https://www.ourcommons.ca/Committees/en/Home}} in our dataset. 

\begin{figure*}[t]
    \centering
    \includegraphics[width=\linewidth, trim=10 210 10 10, clip]{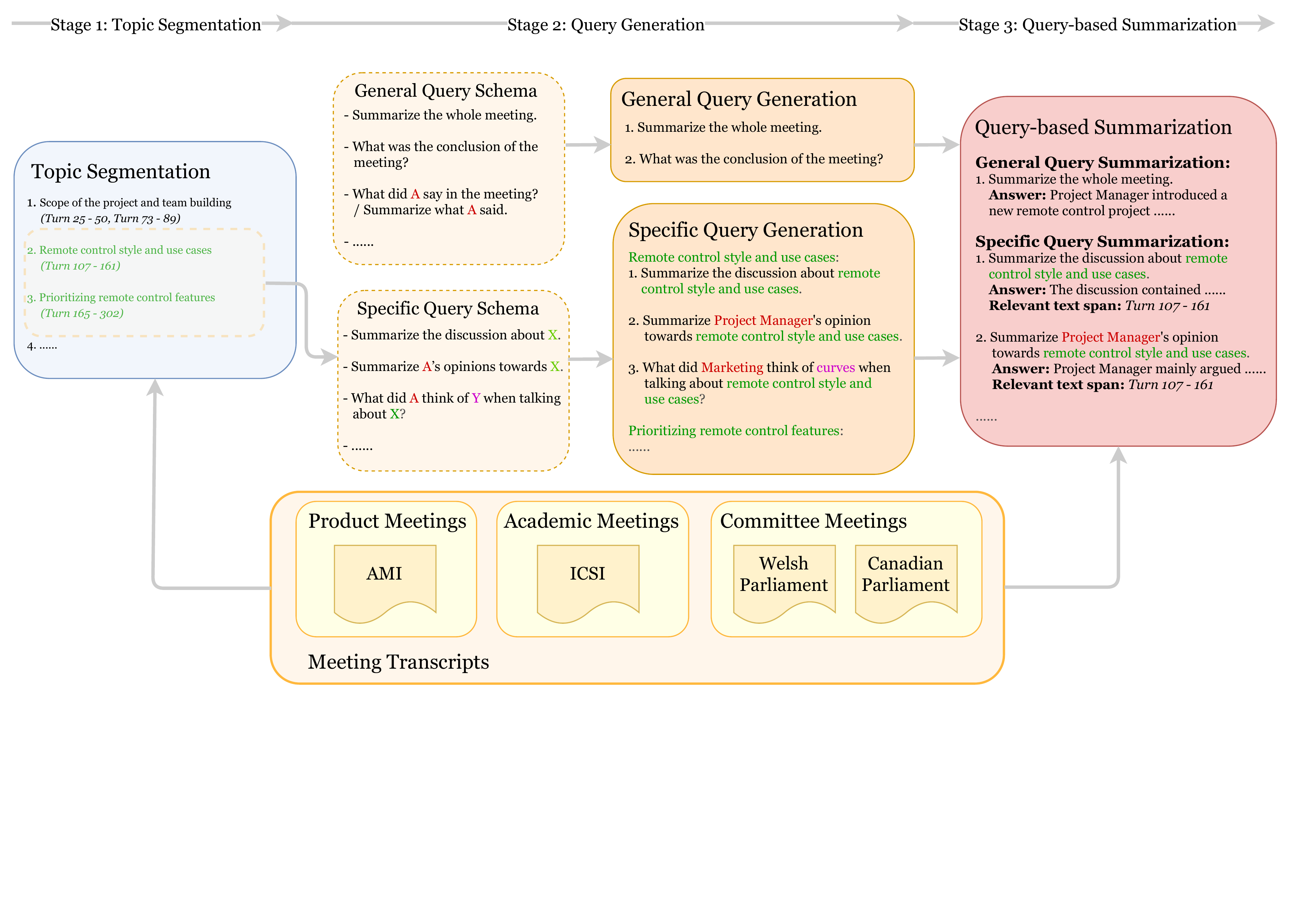}
    \caption{Overall annotation pipeline. It is divided into three stages: Stage 1 is to annotate main topics and their relevant text spans; Stage 2 is to generate queries based on query schema lists; Stage 3 is to annotate the summaries according to the queries. The pipeline was implemented upon the collected meetings of multiple domains.}
    \label{fig:local}
\end{figure*}

\subsection{Annotation Pipeline}
\label{sec:pipeline}
After collecting meeting transcripts, we recruited annotators and required them to annotate by following annotation instruction.
As illustrated in Figure \ref{fig:local}, the annotation process is composed by three stages: topic segmentation, query generation, and query-based summarization. 

\paragraph{Topic Segmentation} Meeting transcripts are usually long and contain discussions about multiple topics. To assist further annotations, we asked annotators to write down the main topics discussed in the meetings, and their relevant text spans, which makes the meeting structure clear. 
As shown in Figure \ref{fig:local}, ``scope of the project and team building'' is one of the annotated main topics, and its relevant text spans of the topic are (\textit{\textbf{Turn}} 25 - 50, \textit{\textbf{Turn}} 73 - 89). More details are listed in Appendix \ref{topic_seg}.

\paragraph{Query Generation} Towards the query-based task, we further asked annotators to design queries by themselves. To cater to the need for multi-granularity contents, we categorized two types of queries: queries related to general information (e.g., the contents of whole meetings, etc.) are called \emph{general queries}; queries focusing on relatively detailed information (e.g., the discussion about certain topics, etc.) are called \emph{specific queries}.

To alleviate the influence of extremely hard queries and focus on the evaluation of query-based summarization capacity, rather than designing queries in an unconstrained way, we asked annotators to generate queries according to the schema. Details of the query schema list are shown in Appendix \ref{sec:list}. The list consists of important facets people might be interested in, including overall contents of discussions, speakers' opinions, the reasons why a speaker proposed an idea, etc., which cover the most common queries over meetings involving multiple people discussing several topics.

To query multi-granularity meeting contents, we further divided the query schema list into general and specific ones, and asked annotators to design queries towards general and specific meeting contents, respectively. In terms of \emph{general query generation}, the annotators were asked to design 1 - 2 general queries according to the general schema list. For \emph{specific query generation}, annotators were asked to first select 2 - 4 main topics and their relevant text spans, and then design around 3 specific queries based on the specific schema list for each main topic. 
To ensure the task to be \emph{summarization} instead of \emph{question answering}, we asked annotators to design queries of which the relevant text spans are more than 10 turns or 200 words. Therefore, our proposed task would differ from \emph{question answering} tasks where models merely need to extract phrases or generate answers based on short text spans, and focus on how to \emph{summarize} based on large stretches of texts. 
Additional details are in Appendix \ref{qg}.

\paragraph{Query-based Summarization} According to the designed queries and meeting transcripts, annotators were asked to do faithful summarization. Being accorded with the meeting transcripts and queries is the most important criterion. We also required annotators to write informative summarization. For example, they could add more details about the reasons why the group/committee made such decisions, and which important ideas the group/committee members proposed, etc. Besides, the annotated summaries should be abstractive, fluent and concise. We set word limits for the answers of general queries (50 - 150 words) and specific queries (20 - 100 words) to keep conciseness. More details are shown in Appendix \ref{query-sum}.

In the end, we organize all the meeting data after accomplishing the three annotation stages. Detailed annotations of one product meeting and one committee meeting are shown in Appendix \ref{qmsum_exp}. Each meeting transcript is accompanied with annotated main topics, queries, their corresponding summaries, and relevant text span information. 

\renewcommand\arraystretch{1.3}
\begin{table*}[t]
    \center \footnotesize
    \tabcolsep0.08 in
    \scalebox{0.99}{
    \begin{tabular}{lccccccc}
    \toprule
    \textbf{Datasets} & \textbf{\# Meetings} & \textbf{\# Turns} & \textbf{\# Len. of Meet.} & \textbf{\# Len. of Sum.} & \textbf{\# Speakers} & \textbf{\# Queries} & \textbf{\# Pairs} \\
    \midrule
    AMI & 137 & 535.6 & 6007.7 & 296.6 & 4.0 & - & 97~/~20~/~20 \\
    ICSI & 59 & 819.0 & 13317.3 & 488.5 & 6.3 & - & 41~/~9~/~9 \\
    \midrule
    Product & 137 & 535.6 & 6007.7 & 70.5 & 4.0 & 7.2 & 690~/~145~/~151 \\
    Academic & 59 & 819.0 & 13317.3 & 53.7 & 6.3 & 6.3 & 259~/~54~/~56 \\
    Committee & 36 & 207.7 & 13761.9 & 80.5 & 34.1 & 12.6 & 308~/~73~/~72 \\
    All & 232 & 556.8 & 9069.8 & 69.6 & 9.2 & 7.8 & 1,257~/~272~/~279 \\
    \bottomrule
    \end{tabular}%
    }
    \caption{Statistics of meeting summarization datasets. The top half of the table is the existing meeting datasets, and the bottom half is the statistics of QMSum. Because a meeting may have multiple queries, \#Pairs here means how many query-summary pairs are contained in the train~/~valid~/~test set.}
  \label{tab:datasets}%
  \vspace{-2mm}
\end{table*}%

\subsection{Additional Details of Annotation Process}
\label{sec:additional}
This section describes how we recruited annotators and how we review the annotations in detail.

\paragraph{Annotator Recruitment} To guarantee annotation quality given the complexity of the task, instead of employing tasks on Amazon Mechanical Turker, we anonymously recruited undergraduate students who are fluent in English. The annotation team consists of 2 native speakers and 10 non-native speakers majoring in English literature.

\paragraph{Annotation Review} To help the annotators fully comprehend the instruction, annotators were trained in a \emph{pre-annotation} process. Annotations were reviewed across all stages in our data collection process by expert of this annotation task. More details of review standards could be found in Appendix \ref{appendix-standards}. 

\subsection{Dataset Statistics and Comparison}
Statistics of the final QMSum dataset is shown in Table \ref{tab:datasets}. 
There are several advantages of QMSum dataset, compared with the previous datasets.

\paragraph{Number of Meetings and Summaries} QMSum includes 232 meetings, which is the largest meeting summarization dataset to our best knowledge.
For each query, there is a manual annotation of corresponding text span in the original meeting, so there are a total of 1,808 question-summary pairs in QMSum. Following the previous work, we randomly select about 15\% of the meetings as the validation set, and another 15\% as the test set.

\paragraph{Briefty} The average length of summaries in QMSum 69.6 is much shorter than that of previous AMI and ICSI datasets. It is because our dataset also focuses on specific contents of the meetings, and the length of their corresponding summaries would not be long. It leaves a challenge about how to precisely capture the related information and compress it into a brief summary. 

\paragraph{Multi-domain Setting} Previous datasets are specified to one domain. However, the model trained on the summarization data of a single domain usually has poor generalization ability~\cite{wang2019exploring, zhong2019closer,DBLP:conf/emnlp/ChenLZDWQH20}. Therefore, QMSum contains meetings across multiple domains: Product, Academic and Committee meetings. We expect that our dataset could provide a venue to evaluate the model's generalization ability on meetings of different domains and help create more robust models.

\section{Method}
In this section, we first define the task of query-based meeting summarization, then describe our two-stage \textit{locate-then-summarize} solution in detail.
\subsection{Problem Formulation}
Existing meeting summarization methods define the task as a sequence-to-sequence problem. Specifically, each meeting transcript $ X = (x_{1}, x_{2}, \cdots, x_{n})$ consists of $n$ turns, and each turn $x_i$ represents the utterance $u_i$ and its speaker $s_i$, that is, $x_i = (u_i, s_i)$. Additionally, each utterance contains $l_i$ words $u_i = (w_1,\cdots, w_{l_i})$. The object is to generate a target summary $Y$ $ = ({y}_{1}, {y}_{2},\cdots,{y}_{m})$ by modeling the conditional distribution $p(y_{1},y_{2},\cdots,y_m|(u_1, s_1),\cdots,(u_n, s_n))$.

However, meetings are usually long conversations involving multiple topics and including important decisions on many different matters, so it is necessary and practical to use queries to summarize a certain part of the meeting. Formally, we introduce a query $Q = (w_1,\cdots, w_{|Q|})$ for meeting summarization task, the objective is to generate a summary $Y$ by modeling $p(y_{1},y_{2},\cdots,y_m|Q, (u_1, s_1),\cdots,(u_n, s_n))$.

\subsection{Locator}
In our two-stage pipeline, the first step requires a model to locate the relevant text spans in the meeting according to the queries, and we call this model a Locator. The reason why we need a Locator here is, most existing abstractive models cannot process long texts such as meeting transcripts. So we need to extract shorter, query-related paragraphs as input to the following Summarizer.

We mainly utilize two methods to instantiate our Locator: Pointer Network~\cite{vinyals2015pointer} and a hierarchical ranking-based model. Pointer Network has achieved widespread success in extractive QA tasks~\cite{DBLP:conf/iclr/Wang017a}. For each question, it will point to the <start, end> pair in the source document, and the span is the predicted answer. Specific to our task, Pointer Network will point to the start turn and the end turn for each query. It is worth noting that one query can correspond to multiple spans in our dataset, so we always extract three spans as the corresponding text for each query when we use Pointer Network as Locator in the experiments.

\begin{figure}[t]
    \centering
    \includegraphics[width=\linewidth, trim=0 100 155 20, clip]{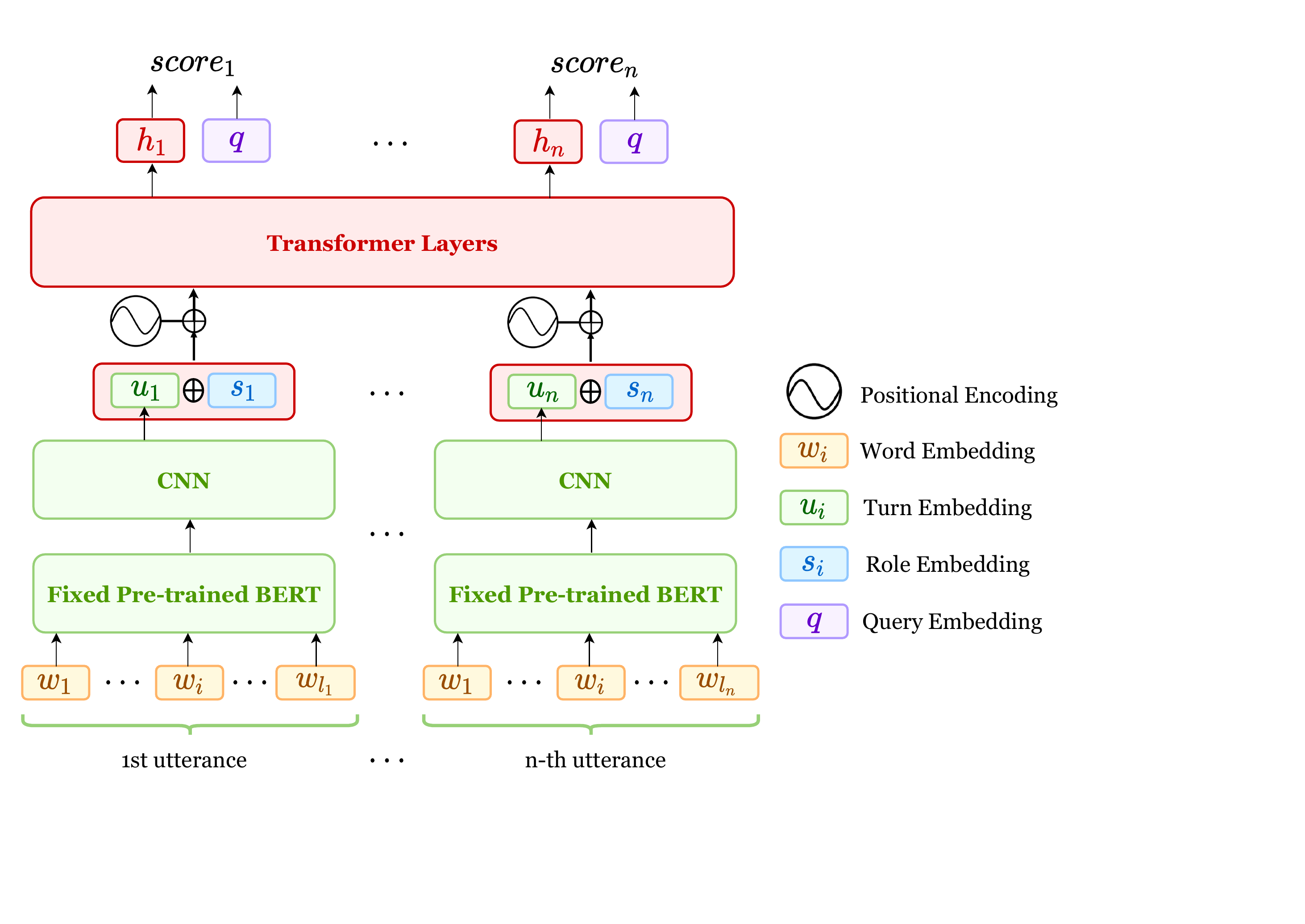}
    \caption{Hierarchical ranking-based locator structure.}
    \label{fig:locator}
    \vspace{-2mm}
\end{figure}

In addition, we design a hierarchical ranking-based model structure as the Locator. As shown in Figure \ref{fig:locator}, we first input the tokens in each turn to a feature-based BERT to obtain the word embedding, where feature-based means we fix the parameters of BERT, so it is actually an embedding layer. Next, CNN \cite{kim2014convolutional} is applied as a turn-level encoder to capture the local features such as bigram, trigram and so on in each turn. Here we do not use Transformer because previous work \cite{kedzie2018content} shows that this component does not matter too much for the final performance. We combine different features to represent the utterance $\mathbf{u}_i$ in each turn, and concatenate the speaker embedding $\mathbf{s}_i$ as the turn-level representation: $\mathbf{x_i} = [\mathbf{u_i}; \mathbf{s_i}]$, where $[;]$ denotes concatenation and $\mathbf{s}_i$ is a vector randomly initialized to represent the speaking style of meeting participants.

Then these turn representations will be contextualized by a document-level Transformer \cite{vaswani2017attention} encoder.

Next, we introduce query embedding $\mathbf{q}$ which is obtained by a CNN (shared parameters with CNN in turn-level encoder) and use MLP to score each turn.

We use binary cross-entropy loss to train our Locator. Finally, turns with the highest scores are selected as the relevant text spans of each query and will be inputted to the subsequent Summarizer. 

\subsection{Summarizer}

Given the relevant paragraphs, our goal in the second stage is to summarize the selected text spans based on the query. We instantiate our Summarizer with the current powerful abstractive models to explore whether the query-based meeting summarization task on our dataset is challenging. To be more specific, we choose the following three models:

\textbf{Pointer-Generator Network} \cite {see2017get} is a popular sequence-to-sequence model with copy mechanism and coverage loss, and it acts as a baseline system in many generation tasks. The input to Pointer-Generator Network (PGNet) is: ``<s> Query </s> Relevant Text Spans </s>''.

\textbf{BART} \cite{DBLP:conf/acl/LewisLGGMLSZ20} is a denoising pre-trained model for language generation, translation and comprehension. It has achieved new state-of-the-art results on many generation tasks, including summarization and abstractive question answering. The input to BART is the same as PGNet.

\textbf{HMNet} \cite{zhu-etal-2020-hierarchical} is the state-of-the-art meeting summarization model. It contains a hierarchical structure to process long meeting transcripts and a role vector to depict the difference among speakers. Besides, a cross-domain pretraining process is also included in this strong model. We add a turn representing the query at the beginning of the meeting as the input of HMNet.

\section{Experiments}
In this section, we introduce the implementation details, effectiveness of Locator, experimental results and multi-domain experiments on QMSum.

\subsection{Implementation Details}
For our ranking-based Locator, the dimension of speaking embedding is 128 and the dimension of turn and query embedding is 512. Notably, we find that removing Transformers in Locator has little impact on performance, so the Locator without Transformer is used in all the experiments. To reduce the burden of the abstractive models, we utilize Locator to extract 1/6 of the original text and input them to Summarizer. The hyperparameters used by PGNet and HMNet are consistent with the original paper. Due to the limitation of computing resources, we use the base version of pre-trained models (including feature-based BERT and BART) in this paper. We use fairseq library\footnote{\url{https://github.com/pytorch/fairseq/tree/master/examples/bart}} to implement BART model. For PGNet and BART, we truncate the input text to 2,048 tokens, and remove the turns whose lengths are less than 5. All results reported in this paper are averages of three runs.

\subsection{Effectiveness of Locator}

\renewcommand\arraystretch{1.3}
\begin{table}[t]
    \centering \footnotesize
    \tabcolsep0.1 in
    \begin{tabular}{lcccc}
        \toprule
        \multicolumn{1}{l}{\multirow{2}[1]{*}{\textbf{Models}}}  & \multicolumn{4}{c}{\textbf{Extracted Length}} \\
         & $\sfrac{1}{6}$ & $\sfrac{1}{5}$ & $\sfrac{1}{4}$ & $\sfrac{1}{3}$ \\
        \midrule
        Random & 58.86 & 63.20 & 67.56 & 73.81 \\
        Similarity & 55.97 & 59.24 & 63.45 & 70.12 \\
        Pointer & 61.27 & 65.84 & 70.13 & 75.96 \\
        Our Locator & \textbf{72.51} & \textbf{75.23} & \textbf{79.08} & \textbf{84.04} \\
        \bottomrule
    \end{tabular}
    \caption{ROUGE-L Recall score between the predicted spans and the gold spans. $\sfrac{1}{6}$ means that the turns extracted by the model account for 1/6 of the original text.}
    \label{tab:Locator}
    \vspace{-2mm}
\end{table}

First, we need to verify the effectiveness of the Locator to ensure that it can extract spans related to the query. Instead of the accuracy of capturing relevant text spans, we focus on the extent of overlap between the selected text spans and the gold relevant text spans. It is because whether the summarization process is built on similar contexts with references or not is essential for Summarizer. Therefore, we use ROUGE-L recall to evaluate the performance of different models under the setting of extracting the same number of turns.

We introduce two additional baselines: Random and Similarity. The former refers to randomly extracting a fixed number of turns from the meeting content, while the latter denotes that we obtain turn embedding and query embedding through a feature-based BERT, and then extract the most similar turns by cosine similarity. As shown in Table \ref{tab:Locator}, because there are usually a large number of repeated conversations in the meetings, Random can get a good ROUGE-L recall score, which can be used as a baseline to measure the performance of the model. Similarity performs badly, 
even worse than Random, which may be due to the great difference in style between the BERT pre-trained corpus and meeting transcripts. Pointer Network is only slightly better than Random. We think this is because in the text of with an average of more than 500 turns, only three <start, end> pairs are given as supervision signals, which is not very informative and therefore is not conducive to model learning.

On the contrary, our hierarchical ranking-based Locator always greatly exceeds the random score, which demonstrates that it can indeed extract more relevant spans in the meeting. Even if 1/6 of the original text is extracted, it can reach a 72.51 ROUGE-L recall score, which significantly reduces the burden of subsequent Summarizer processing long text while ensuring the amount of information.

\subsection{Experimental Results on QMSum}

\renewcommand\arraystretch{1.3}
\begin{table}[t]
    \centering \footnotesize
    \tabcolsep0.15 in
    \begin{tabular}{lccc}
        \toprule
        \textbf{Models}  & \textbf{R-1} & \textbf{R-2}  & \textbf{R-L}\\
        \midrule
        Random & 12.03 & 1.32 & 11.76 \\
        Ext. Oracle & 42.84 & 16.86 & 39.20 \\
        \midrule
        TextRank & 16.27 & 2.69 & 15.41 \\
        PGNet & 28.74 & 5.98 & 25.13 \\
        BART & 29.20 & 6.37 & 25.49 \\
        PGNet$^*$  & 31.37 & 8.47 & 27.08 \\
        BART$^*$ & 31.74 & 8.53 & \textbf{28.21} \\
        HMNet$^*$ & \textbf{32.29} & \textbf{8.67} & 28.17 \\
        \midrule
        PGNet$^\dag$ & 31.52 & 8.69 & 27.63 \\
        BART$^\dag$ & 32.18 & 8.48 & 28.56 \\
        HMNet$^\dag$ & \textbf{36.06} & \textbf{11.36} & \textbf{31.27} \\
        \bottomrule
    \end{tabular}
    \caption{Experimental results on QMSum dataset. We use standard ROUGE F-1 score to evaluate different models. The models with $^*$ denotes they use the spans extracted by our Locator as the input and $^\dag$ indicates this Summarizer uses gold spans as the input.}
    \label{tab:Summarizer}
    \vspace{-3mm}
\end{table}

\renewcommand\arraystretch{1.3}
\begin{table*}[t]
    \center \footnotesize
    \tabcolsep0.1 in
    \begin{tabular}{lcccccccccccc}
    \toprule
    \multicolumn{1}{l}{\multirow{2}[1]{*}{\textbf{Datasets}}} &
    \multicolumn{3}{c}{\textbf{Product}} &
    \multicolumn{3}{c}{\textbf{Academic}} &
    \multicolumn{3}{c}{\textbf{Committee}} &
    \multicolumn{3}{c}{\textbf{All}} \\
     & R-1 & R-2 & R-L & R-1 & R-2 & R-L & R-1 & R-2 & R-L & R-1 & R-2 & R-L \\
    \midrule
    \textbf{Pro.} & \cellcolor[rgb]{ .851,  .851,  .851}35.43 & \cellcolor[rgb]{ .851,  .851,  .851}10.99 & \cellcolor[rgb]{ .851,  .851,  .851}31.37 & 22.59 & 3.41 & 19.82 & 24.48 & 3.84 & 21.94 & 30.02 & 7.58 & 26.62 \\
    \textbf{Aca.} & 27.19 & 4.86 & 24.09 & \cellcolor[rgb]{ .851,  .851,  .851}26.69 & 4.32 & 22.58 & 27.84 & 4.29 & 25.10 & 27.22 & 4.59 & 24.02 \\
    \textbf{Com.} & 25.56 & 3.48 & 22.17 & 23.91 & 2.99 & 20.23 & \cellcolor[rgb]{ .851,  .851,  .851}32.52 & \cellcolor[rgb]{ .851,  .851,  .851}6.98 & \cellcolor[rgb]{ .851,  .851,  .851}27.71 & 27.07 & 4.28 & 23.21 \\
    \textbf{All} & 34.93 & 10.78 & 31.21 & 26.47 & \cellcolor[rgb]{ .851,  .851,  .851}5.05 & \cellcolor[rgb]{ .851,  .851,  .851}23.01 & 31.16 & 6.47 & 27.52 & \cellcolor[rgb]{ .851,  .851,  .851}32.18 & \cellcolor[rgb]{ .851,  .851,  .851}8.48 & \cellcolor[rgb]{ .851,  .851,  .851}28.56 \\
    \bottomrule
    \end{tabular}%
    \caption{Multi-domain and cross-domain summarization experiments. Each row represents the training set, and each column represents the test set. The gray cells denote the best result on the dataset in this column.We use BART$^\dag$ for these experiments and use standard ROUGE F-1 score to evaluate the model performance.}
  \label{tab:cross_domain}%
  \vspace{-2mm}
\end{table*}%

For comparison, we introduce two basic baselines: Random and Extractive Oracle. We randomly sample 10 turns of the original meeting for each query as an answer and this is the Random baseline in Table \ref{tab:Summarizer}. Besides, we implement the Extractive Oracle, which is a greedy algorithm for extracting the highest-scoring sentences, usually regarded as the the upper bound of the extractive method \cite{nallapati2017summarunner}. An unsupervised method, TextRank is also included in our experiment. We treat each turn as a node and add a query node to fully connect all nodes. Finally, the 10 turns with the highest scores are selected as the summary.

Table \ref{tab:Summarizer} shows that the performance of three typical neural network models is significantly better than Random and TextRank. When equipped with our Locator, both PGNet and BART have brought evident performance improvements (PGNet: 28.74 -> 31.37 R-1, BART: 29.20 -> 31.74 R-1). Compared to PGNet$^*$, the advantage of BART$^*$ lies in the ROUGE-L score (1.13 improvement), which indicates that it can generate more fluent sentences. The current state-of-the-art meeting summarization model HMNet achieves the best performance, which may be attributed to its cross-domain pre-training process making HMNet more familiar with the style of meeting transcripts.

In addition, we also use the gold text spans as the input of different models to measure the performance loss caused by Locator. Surprisingly, for models (PGNet and BART) that need to truncate the input text, although Locator is an approximate solution, the models equipped with it can achieve comparable results with the models based on gold span inputs. Therefore, in this case, our two-stage pipeline is a simple but effective method in the meeting domain. However, for some models (HMNet) that use a hierarchical structure to process long text, inputting gold text spans can still bring 
huge performance improvements.

\subsection{Experiments on Different Domains}

In addition, we also conduct multi-domain and cross-domain experiments. First, we perform in-domain and out-domain tests in the three domains of QMSum dataset. In Table \ref{tab:cross_domain}, we can conclude that there are obvious differences between these three domains. For instance, the models trained on the Academic and Committee domains perform poorly when tested directly on the Product domain, with only the ROUGE-L scores of 24.09 and 22.17 respectively. However, the model trained on the single domain of Product can achieve a ROUGE-L score of 31.37, which illustrates although these domains are all in the form of meeting transcript, they still have visible domain bias.

On the other hand, when we train all the domains together, we can obtain a robust summarization model. Compared with models trained on a single domain, models trained on QMSum can always achieve comparable results. In the Academic domain, the model with multi-domain training can even get higher ROUGE-2 (5.05 vs 4.32) and ROUGE-L (23.01 vs 22.58) scores. These results show that the multi-domain setting in meeting summarization task is apparently necessary and meaningful. Meeting transcripts cover various fields, making the transfer of models particularly difficult. Therefore, we need to introduce multi-domain training to make the model more robust, so it can be applied to more practical scenarios.

\section{Analysis}

In this section, we conduct comprehensive analysis of query types and errors in the model output.

\subsection{Analysis of Query Types}

We manually divide the query in QMSum into five aspects: personal opinion, multi-person interaction, conclusion or decision, reason, and overall content. For example, ``Summarize the whole meeting.'' requires a summary of the overall content and ``Why did A disagree with B?'' requires a summary of some reasons. The questions we are concerned about are: \textit{what is the distribution of different types of queries in QMSum? Are there differences in the difficulty of different types of queries?}

\renewcommand\arraystretch{1.3}
\begin{table}[t]
    \centering \footnotesize
    \tabcolsep0.06 in
    \begin{tabular}{lcccccc}
        \toprule
          & \textbf{Opin.} & \textbf{Inter.} & \textbf{Con./Dec.} & \textbf{Reason} & \textbf{Overall} \\
        \midrule
        Num. & 22 & 40 & 19 & 7 & 12 \\
        Diff. 1 & 2.2 & 1.6 & 1.7 & 2.4 & 2.0 \\
        Diff. 2 & 1.9 & 2.1 & 2.3 & 2.2 & 1.6 \\
        R-L & 27.0 & 30.1 & 26.1 & 24.9 & 30.9 \\
        \bottomrule
    \end{tabular}
    \caption{The number, human evaluation and model performance of different types of queries. Diff. 1 represents the difficulty of locating relevant information and Diff. 2 represents the difficulty of organizing content.}
    \label{tab:query_type}
    \vspace{-2mm}
\end{table}

To figure out the above issues, we randomly sample 100 queries from the test set, count the number of each type, and score the difficulty of each query. Table \ref{tab:query_type} illustrates that answering 40\% of queries requires summarizing the interaction of multiple people, and the queries that focus on personal opinions and different aspects of conclusions or decisions account for almost 20\% each. Besides, queries about a specific reason are less frequent in the meetings.

We also perform a human evaluation of the difficulty of various query types. For each query, the relevant text spans and query-summary pair are shown to annotators. Annotators are asked to score the difficulty of this query in two dimensions: 1) the difficulty of locating relevant information in the original text; 2) the difficulty of organizing content to form a summary. For each dimension, they can choose an integer between 1 and 3 as the score, where 1 means easy and 3 means difficult.

As we can see from Table \ref{tab:query_type}, query about reasons is the most difficult to locate key information in related paragraphs, and this type of query is also challenging to organize and summarize reasonably. Queries about multi-person interaction and overall content are relatively easy under human evaluation scores. The relevant paragraphs of the former contain multi-person conversations, which are usually redundant, so the effective information is easier to find; the latter only needs to organize the statements in the chronological order of the meeting to write a summary, so it has the lowest Diff. 2 score. The model performance also confirms this point, BART can get more than 30 R-L score on these two types of queries, but performs poorly on the rest. Therefore, the remaining three types of queries in QMSum are still very challenging even for powerful pre-trained models, and further research is urgently needed to change this situation.

\subsection{Error Analysis}
Although ROUGE score can measure the degree of overlap between the generated summary and the gold summary, it cannot reflect the factual consistency between them or the relevance between the predicted summary and the query. Therefore, in order to better understand the model performance and the difficulty of the proposed task, we sample 100 generated summaries for error analysis. Specifically, we ask 10 graduate students to do error analysis on the sampled summaries. Each summary is viewed by two people. They discuss and agree on whether the sample is consistent with the original facts and whether it is related to the query.

According to \citet{DBLP:conf/aaai/CaoWLL18}, nearly 30\% of summaries generated by strong neural models contain factual errors. This problem is even more serious on QMSum: we find inconsistent facts in 74\% of the samples, which may be because the existing models are not good at generating multi-granularity summaries. Although BART can achieve state-of-the-art performance in the single-document summarization task, it does not seem to be able to truly understand the different aspects of the meeting, thus create factual errors. What's worse, 31\% summaries are completely unrelated to the given query. This not only encourages us to design more powerful models or introduce more prior knowledge to overcome this challenge, but also shows better metrics are needed to evaluate model performance in generating multi-granularity summaries.

\section{Conclusion}
We propose a new benchmark, QMSum, for query-based meeting summarization task.
We build a \textit{locate-then-summarize} pipeline as a baseline and further investigate variants of our model with different Locators and Summarizers, adopt different training settings including cross-domain and multi-domain experiments to evaluate generalizability, and analyze the task difficulty with respect to query types. The new task and benchmark leave several open research directions to explore: 1) how to process the long meeting discourses;
2) how to make a meeting summarization model generalize well; 
3) how to generate summaries consistent with both meeting transcripts and queries.
4) how to reduce the annotation cost for meeting summarization. 

\section*{Acknowledgements}
The Language, Information, and Learning lab at Yale (LILY lab) would like to acknowledge the research grant from Microsoft Research. We would also like to thank annotators for their hard work and reviewers for their valuable comments. 

\section*{Ethics Consideration}
\label{ethics}
We propose a novel query-based meeting summarization task, accompanying with a high-quality dataset QMSum. Since the paper involves a new dataset and NLP application, this section is further divided into the following two parts.

\subsection{New Dataset}
\paragraph{Intellectual Property and Privacy Rights}
Collecting user data touches on the intellectual property and privacy rights of the original authors: both of the collected meeting transcripts and recruited annotators. We ensure that the dataset construction process is consistent with the intellectual property and privacy rights of the original authors of the meetings. All the meeting transcripts we collected are public and open to use according to the regulation~\footnote{\url{http://groups.inf.ed.ac.uk/ami/corpus/license.shtml}}~\footnote{\url{http://groups.inf.ed.ac.uk/ami/icsi/license.shtml}}~\footnote{\url{https://senedd.wales/en/help/our-information/Pages/Open-data.aspx}}~\footnote{\url{https://www.ourcommons.ca/en/important-notices}}. The annotation process is consistent with the intellectual property and privacy rights of the recruited annotators as well.

\paragraph{Compensation for Annotators}
We estimated the time for annotating one meeting is around 1 - 2 hours. Therefore, we paid annotators around \$14 for each product and academic meeting and \$28 for each committee meeting. To further encourage annotators to work on annotations, we proposed bonus mechanism: the bonus of each of the 5th to 8th meetings would be \$4; the bonus of each of the 9th to 12th meetings would be \$5, and so on. Some of the authors also did annotations and they were paid as well.

\paragraph{Steps Taken to Avoid Potential Problems}
The most possible problems which may exist in the dataset is bias problem and the inconsistency among queries, annotated summaries and original meeting contents. With regard to bias problem, we find that product meeting dataset rarely contains any explicit gender information, but annotators still tended to use `he' as pronoun. To avoid the gender bias caused by the usage of pronouns, we required annotators to replace pronouns with speaker information like `Project Manager', `Marketing' to avoid the problem. Also, when designing queries based on query schema list, we found that annotators usually used the same query schema, which might lead to bias towards a certain type of query. Therefore, we asked the annotators to use different schemas as much as possible. For the inconsistency problem, each annotation step was strictly under supervision by `experts' which are good at annotation and could be responsible for reviewing.

\subsection{NLP Applications}
\paragraph{Intended Use}
The query-based meeting summarization application is aiming at summarizing meetings according to queries from users. We could foresee that the trained model could be applied in companies to further improve the efficiency of workers, and help the staff comprehensively understand the meeting contents. The annotated QMSum dataset could be used as a benchmark for researchers to study how to improve the performance of summarization on such long texts and how to make models more generalizable on the meetings of different domains.

\paragraph{Failure Mode}
The current baseline models still tend to generate ungrammatical and factually inconsistent summaries. If a trained baseline model was directly applied in companies, the misinformation would negatively affect comprehension and further decision making. Further efforts are needed to generate high-quality summaries which are fluent and faithful to the meeting transcripts and queries.

\paragraph{Bias}
Training and test data are often biased in ways that limit system accuracy on domains with small data size or new domains, potentially causing distribution mismatch issues.
In the data collection process, we control for the gender bias caused by pronouns such as `he' and `she' as much as possible. Also, we attempt to control the bias towards a certain type of query schema by requiring annotators to use diverse schemas as much as possible. However, we admit that there might be other types of bias, such as political bias in committee meetings. Thus, the summarization models trained on the dataset might be biased as well. and We will include warnings in our dataset.

\paragraph{Misuse Potential}
We emphasize that the application should be used with careful consideration, since the generated summaries are not reliable enough. It is necessary for researchers to develop better models to improve the quality of summaries. Besides, if the model is trained on internal meeting data, with the consideration of intellectual property and privacy rights, the trained model should be used under strict supervision. 

\paragraph{Collecting Data from Users}
Future projects have to be aware of the fact that some meeting transcripts are intended for internal use only. Thus, researchers should be aware of the privacy issues about meeting data before training the model.

\bibliography{anthology,custom}
\bibliographystyle{acl_natbib}

\clearpage

\appendix

\section{Appendix}

\subsection{Query Schema List}
\label{sec:list}
As mentioned in Section~\ref{sec:pipeline}, we make query schema list to help annotators design queries. Towards general and specific meeting contents, we further divide the query schema list into \emph{general query schema list} and \emph{specific query schema list}. The detailed list is shown in Table~\ref{tab:schema_list}.

\subsection{Other Details of Annotation Instruction}
\label{others}
We show other details except those in Section \ref{data_construction}.

\subsubsection{Topic Segmentation}
\label{topic_seg}
\paragraph{Topics.}
We require annotators to use noun phrases to represent the main topics. As we hope to select the most important topics, the number of annotated main topics should not be too much, and 3 to 8 is proper range.

\paragraph{Distribution of Relevant Text Spans in Meeting Transcripts.}
Relevant text spans of a main topic may be scattered in different parts of meetings, e.g., the main topic `scope of the project and team building' in the leftmost part of Figure \ref{fig:local}. So annotators are asked to label \emph{all} the relevant spans, and these spans are allowed to be not contiguous. 

\paragraph{Whether Chatting Belongs to an Independent Main Topic or not.} 
Main topics should be objectively cover most of the meeting contents. In meetings, group members might take much time chatting. Though this is not close to the main theme of the meetings, it should also be counted as a main topic if the chat took lots of time.

\subsubsection{Query Generation}
\label{qg}
\paragraph{Diverse Query Types.}
For the specific queries, we encourage the annotators to choose different schemas to compose queries, since we intend to diversify the query types and reduce the bias towards certain query types.

\paragraph{Independent Query-answer Pairs.} 
Each query-answer pair should be independent, that is, not dependent on previous queries and answers. For example, for the query `How did they reach agreement afterwards?', it seems that the query is dependent on the previous annotations, and it is ambiguous to know when they reached agreement and what the agreement referred to if we treat the query independently. Annotators should specify the information of when they reached an agreement and what the agreement is to make the query clear. For example, annotators could rewrite the query as `How did the group agree on \textcolor{subtopic}{the button design} when discussing \textcolor{topic}{the design of remote control}?'.

\subsubsection{Query-based Summarization}
\label{query-sum}
\paragraph{Informative Writing.}
When answering queries like ‘What did \textcolor{speaker}{A} think of \textcolor{topic}{X}?’, annotators were required to not only summarize what \textcolor{speaker}{A} said, but also briefly mention the context which is relevant to what \textcolor{speaker}{A} said. This is designed to rich the contents of summaries and further challenge the model's capability of summarizing relevant contexts.

\paragraph{Relevant text spans' annotation.}
Since there might be multiple relevant text spans, we asked annotators to annotate all of them.
\paragraph{The Usage of Tense.} 
Since all the meetings happened, we ask annotators to use past tense.
\paragraph{How to Denote Speakers.} If the gender information is unclear, we would ask annotators not to use ‘he/she’ to denote speakers. We also asked them not to abbreviations (e.g., PM) to denote speakers, and use the full name like ‘Project Manager’ instead.
\paragraph{Abbreviations.} In the raw meeting transcripts, some of abbreviations are along with character `\_'. If the annotators encountered abbreviations, e.g., `L\_C\_D\_', `A\_A\_A\_', etc., they would be required to rewrite them like LCD or AAA.

\subsection{Annotation Review Standards}
\label{appendix-standards}
We launched a \emph{`pre-annotation'} stage in which annotators were asked to try annotating one meeting and the \emph{experts} who are good at our annotation task would review it and instruct the annotators by providing detailed feedback. Requirements include 1) faithfulness, 2) informativeness, 3) lengths of the relevant text spans of designed queries, 4) typo errors, etc. They could continue annotating only if they passed \emph{`pre-annotation'} stage. After \emph{`pre-annotation'}, \emph{experts} would keep carefully reviewing all the annotations according to the requirements. We write down the standards for the three annotation stages individually in annotation instruction. Details of annotation review standards could be referred to Appendix \ref{others}.

\subsection{Examples of QMSum Dataset}
\label{qmsum_exp}
We show two examples of our proposed QMSum dataset. One belongs to product meeting (Table \ref{tab:group_meeting_exp}), and the other one is about committee meeting (Table \ref{tab:committee_meeting_exp}).

\begin{table*}[ht]
\centering
\begin{tabular}{l}
\toprule
\textbf{General Query Schema List}  \\
\small \textbf{\textcolor{speaker}{A}: Speaker}  \\ \midrule
\small 1. Summarize the whole meeting. / What did the group/committee discuss in the meeting? (Mandatory) \\
\small 2. What did \textcolor{speaker}{A} say in the meeting? / Summarize what \textcolor{speaker}{A} said. \\
\small 3. What was the conclusion / decision of the meeting? \\
\small 4. What was the purpose of the meeting? \\
\small 5. How did the group/committee split the work? \\ \midrule
\textbf{Specific Query Schema List} \\
\small \textbf{\textcolor{speaker}{A, B}: Speakers, \textcolor{topic}{X}: Annotated Main Topics, \textcolor{subtopic}{Y}: Subtopics regarding to \textcolor{topic}{X}} \\ \midrule
\small 1. Summarize the discussion about \textcolor{topic}{X}. / What did the group/committee discuss \textcolor{topic}{X}? (Mandatory) \\
\small 2. Why did the group/committee decide to do sth. when discussing \textcolor{topic}{X}? \\
\small 3. What did \textcolor{speaker}{A} think of \textcolor{subtopic}{Y} when discussing \textcolor{topic}{X}? / Summarize \textcolor{speaker}{A}’s opinions towards \textcolor{subtopic}{Y}. \\
\small 4. What did \textcolor{speaker}{A} think of \textcolor{topic}{X}? / Summarize \textcolor{speaker}{A}’s opinions towards \textcolor{topic}{X}. / What did A propose in the discussion about \textcolor{topic}{X}? \\
\small 5. What was the advantage / disadvantage of sth. with regard to \textcolor{topic}{X}? \\
\small 6. Why did \textcolor{speaker}{A} think regarding to \textcolor{topic}{X}? \\
\small 7. Why did \textcolor{speaker}{A} agree / disagree with certain ideas? / Provide the reasons why \textcolor{speaker}{A} held certain opinions towards \textcolor{topic}{X}. \\
\small 8. Why did \textcolor{speaker}{A} think of \textcolor{subtopic}{Y} when discussing \textcolor{topic}{X}? \\
\small 9. What was the decision / conclusion of the discussion about \textcolor{topic}{X}? / Summarize the decision of the discussion about \textcolor{topic}{X}. \\
\small 10. Why did \textcolor{speaker}{A} agree / disagree with \textcolor{speaker}{B} when discussing \textcolor{topic}{X}? \\
\small 11. What did \textcolor{speaker}{A} recommend to do when discussing \textcolor{topic}{X} and why? \\
\small 12. What did \textcolor{speaker}{A} learn about topic \textcolor{topic}{X}? \\
\small 13. What did \textcolor{speaker}{A} and \textcolor{speaker}{B} discuss \textcolor{topic}{X}? \\ \bottomrule
\end{tabular}
\caption{General and specific query schema lists. \textcolor{speaker}{A, B} denote speaker names. \textcolor{topic}{X} indicates one of the annotated main topics, and \textcolor{subtopic}{Y} means the subtopics regarding to \textcolor{topic}{X}.}
\label{tab:schema_list}
\end{table*}

\begin{table*}[ht]
\normalsize
\centering
\scalebox{0.86}{
\begin{tabular}{l}
\toprule
\small \textbf{Product Meeting (IS1000a)}  \\
\small \textbf{Color: \textcolor{speaker}{Speakers}, \textcolor{topic}{Main Topics}, \textcolor{subtopic}{Subtopics}} \\ 
\midrule
\small \textit{\textbf{Turn 0}}: \textcolor{speaker}{User Interface Designer}: Okay. \\
\small ...... ...... \\
\small \textit{\textbf{Turn 243}}: \textcolor{speaker}{Project Manager}: Well, this uh this tool seemed to work. \\
\small ...... ...... \\
\small \textit{\textbf{Turn 257}}: \textcolor{speaker}{Project Manager}: More interesting for our company of course, uh profit aim, about fifty million Euro. So we have \\
\small to sell uh quite a lot of this uh things. ...... \\
\small \textit{\textbf{Turn 258}}: \textcolor{speaker}{User Interface Designer}: Ah yeah, the sale man, four million. \\
\small \textit{\textbf{Turn 259}}: \textcolor{speaker}{User Interface Designer}: Maybe some uh Asian countries. Um also important for you all is um the production cost \\ 
\small must be maximal uh twelve uh twelve Euro and fifty cents. \\
\small ...... ...... \\
\small \textit{\textbf{Turn 275}}: \textcolor{speaker}{Project Manager}: So uh well I think when we are working on the international market , uh in principle it has enou- \\
\small gh customers. \\
\small \textit{\textbf{Turn 276}}: \textcolor{speaker}{Industrial Designer}: Yeah. \\
\small \textit{\textbf{Turn 277}}: \textcolor{speaker}{Project Manager}: Uh so when we have a good product we uh we could uh meet this this aim, I think. So, that about \\
\small finance. And uh now just let have some discussion about what is a good remote control and uh well keep in mind this this first \\
\small point, it has to be original, it has to be trendy, it has to be user friendly. ...... \\
\small ...... ...... \\
\small \textit{\textbf{Turn 400}}: \textcolor{speaker}{Project Manager}: Keep it in mind. \\ \midrule
\small \textbf{Annotated Main Topics} \\ \midrule
\small Scope of the project and team building (\textit{\textbf{Turn 41 - 245}}) \\ 
\small Cost constraints and financial targets of the new remote control project (\textit{\textbf{Turn 246 - 277}}) \\
\small Remote control style and use cases (\textit{\textbf{Turn 277 - 295}}) \\
\small Prioritizing remote control features (\textit{\textbf{Turn 343 - 390}}) \\ \midrule
\small \textbf{Queries and Annotated Summaries} \\ \midrule
\small \textit{\textbf{Query 1}}: Summarize the whole meeting. \\
\small \textit{\textbf{Answer}}: Project Manager introduced a new remote control project for television sets, and the team got acquainted with each \\
\small other and technical devices. The remote control would be priced at 25 Euros and a production cost of 12.5 Euros. ...... \\
......\\
\small \textit{\textbf{Query 2}}: What did the group discuss about \textcolor{topic}{prioritizing remote control features}? \\
\small \textit{\textbf{Answer}}: User Interface Designer and Industrial Designer expressed a desire to integrate cutting-edge features into the remote. \\
\small Marketing pointed out that most of the market would buy it for standard use, like changing channels and adjusting volume ...... \\
\small \textit{\textbf{Relevant Text Spans}}: \textit{\textbf{Turn 343 - 390}} \\
......\\
\small \textit{\textbf{Query 4}}: Why did \textcolor{speaker}{Marketing} disagree with \textcolor{speaker}{Industrial Designer} when discussing \textcolor{topic}{prioritizing remote control features}? \\
\small \textit{\textbf{Answer}}: Marketing believed that fancy features like IP would not be used by most people. The overwhelming majority of us-\\
\small ers would want convenient channel browsing and volume adjustment features ...... \\
\small \textit{\textbf{Relevant Text Spans}}: \textit{\textbf{Turn 358}} \\
......\\
\small \textit{\textbf{Query 7}}: What did \textcolor{speaker}{Project Manager} think of \textcolor{topic}{the cost constraints and financial targets of the new remote control project}? \\
\small \textit{\textbf{Answer}}: Project Manager introduced the financial information: 25 Euro selling price and 12.5 Euro production cost. Project \\
\small Manager then went on to elaborate that the target market would primarily consist of Europe and North America. ...... \\
\small \textit{\textbf{Relevant Text Spans}}: \textit{\textbf{Turn 248 - 277}} \\
\bottomrule
\end{tabular}
}
\caption{A product meeting annotation example in QMSum dataset.}
\label{tab:group_meeting_exp}
\end{table*}

\begin{table*}[ht]
\normalsize
\centering
\scalebox{0.91}{
\begin{tabular}{l}
\toprule
\small \textbf{Committee Meeting (Education 4)}  \\
\small \textbf{Color: \textcolor{speaker}{Speakers}, \textcolor{topic}{Main Topics}, \textcolor{subtopic}{Subtopics}} \\ 
\midrule
\small \textit{\textbf{Turn 0}}: \textcolor{speaker}{Lynne Neagle AM}: Okay, good morning, everyone. Welcome to the Children, Young People and Education Commit- \\ 
\small tee this morning. I've received apologies for absence from ...... \\
\small ...... ...... \\
\small \textit{\textbf{Turn 31}}: \textcolor{speaker}{David Hopkins}: Yes, sure. The delegation levels are already very high in most authority areas, and we've got agreeme- \\
\small nts in place with the Government to make sure that more money, or as much money as possible ...... \\
\small \textit{\textbf{Turn 32}}: \textcolor{speaker}{Sian Gwenllian AM}: Okay. But just the pressures coming in with the new Act et cetera could mean more expulsions. \\
\small \textit{\textbf{Turn 33}}: \textcolor{speaker}{David Hopkins}: It shouldn't, but it could. It's difficult to know how headteachers and governing bodies will react. ...... \\
\small ...... ...... \\
\small \textit{\textbf{Turn 44}}: \textcolor{speaker}{Sharon Davies}: As Nick said, it does get more difficult at key stage 4, and it's working, then, with. It comes back to \\
\small that team-around-the-family approach ...... \\
\small ...... ...... \\
\small \textit{\textbf{Turn 47}}: \textcolor{speaker}{David Hopkins}: I don't think I'm allowed to say at this point. \\
\small ...... ...... \\
\small \textit{\textbf{Turn 228}}: \textcolor{speaker}{Lynne Neagle AM}: Item 4, then, is papers to note. Just one paper today, which is the Welsh Government's respon- \\
\small se to the committee's report on the scrutiny of the Welsh Government's draft budget 2020-1. ...... \\ \midrule
\small \textbf{Annotated Main Topics} \\ \midrule
\small An increase of exclusions from school and current measures against it (\textit{\textbf{Turn 1 - 19}}, \textit{\textbf{Turn 158 - 172}}) \\ 
\small The funding issues (\textit{\textbf{Turn 20 - 38}}, \textit{\textbf{Turn 177 - 179}}) \\
\small The networking within the PRU and the transition arrangements (\textit{\textbf{Turn 39 - 56}}) \\
\small ...... \\ 
\small Schools' awareness of early trauma ACEs (\textit{\textbf{Turn 180 - 188}}) \\ \midrule
\small \textbf{Queries and Annotated Summaries} \\ \midrule
\small \textit{\textbf{Query 1}}: Summarize the whole meeting. \\
\small \textit{\textbf{Answer}}: The meeting was mainly about the reasons behind and the measurements against the increasing exclusions from school. \\
\small The increase brought more pressure to EOTAS in the aspects of finance, transition, curriculum arrangement and the recruitment of \\
\small professional staff. Although much time and finance had been devoted to the PRU ...... \\
......\\
\small \textit{\textbf{Query 4}}: What was considered by \textcolor{speaker}{David Hopkins} as \textcolor{subtopic}{the factor that affected exclusions}? \\
\small \textit{\textbf{Answer}}: David Hopkins did not think that the delegation levels were not high enough in most authority areas. Instead, he thought \\
\small they had got agreements with the government to make sure that enough money was devolved to school. The true decisive factor w- \\
\small as the narrow measure at the end of Stage 4 that drove the headmasters to exclude students or put them into another school. \\
\small \textit{\textbf{Relevant Text Spans}}: \textit{\textbf{Turn 31 - 33}} \\
......\\
\small \textit{\textbf{Query 6}}: What was the major challenge of \textcolor{subtopic}{the transition of the excluded students}? \\
\small \textit{\textbf{Answer}}: The students coming to the end of their statutory education were facing the biggest challenge, for it would be far more di- \\
\small fficult for them to go back into the mainstream education process when they turned 15 or 16, not to mention the transition into fur- \\
\small ther education, such as colleges. \\
\small \textit{\textbf{Relevant Text Spans}}: \textit{\textbf{Turn 44 - 49}} \\
......\\
\bottomrule
\end{tabular}
}
\caption{A committee meeting annotation example in QMSum dataset.}
\label{tab:committee_meeting_exp}
\end{table*}

\end{document}